# Dynamic Operating System Scheduling Using Double DQN: A Reinforcement Learning Approach to Task Optimization

Xiaoxuan Sun
Independent Researcher
Mountain View, USA

Yifei Duan
University of Pennsylvania
Philadelphia, USA

Yingnan Deng
Georgia Institute of Technology
Atlanta, USA

Fan Guo
Illinois Institute of Technology
Chicago, USA

Guohui Cai
Illinois Institute of Technology
Chicago, USA

Yuting Peng*
New York University
New York, USA

*Abstract-In this paper, an operating system scheduling algorithm based on Double DQN (Double Deep Q network) is proposed, and its performance under different task types and system loads is verified by experiments. Compared with the traditional scheduling algorithm, the algorithm based on Double DQN can dynamically adjust the task priority and resource allocation strategy, thus improving the task completion efficiency, system throughput, and response speed. The experimental results show that the Double DQN algorithm has high scheduling performance under light load, medium load and heavy load scenarios, especially when dealing with I/O intensive tasks, and can effectively reduce task completion time and system response time. In addition, the algorithm also shows high optimization ability in resource utilization and can intelligently adjust resource allocation according to the system state, avoiding resource waste and excessive load. Future studies will further explore the application of the algorithm in more complex systems, especially scheduling optimization in cloud computing and large-scale distributed environments, combining factors such as network latency and energy efficiency to improve the overall performance and adaptability of the algorithm.*

## I. INTRODUCTION

In modern computing environments, the operating system serves as the critical intermediary between computer hardware and applications. The efficiency of its scheduling algorithms plays a crucial role in system performance. The primary goal of operating system scheduling algorithms is to allocate computing resources (such as processor time, memory, and I/O devices) fairly and effectively, ensuring that processes can execute efficiently. This, in turn, improves system response time, throughput, and user experience [1]. With the development of computer hardware, particularly the widespread use of multi-core processors and large-scale parallel computing, traditional scheduling methods are no longer sufficient to meet the increasingly complex demands of modern applications. Therefore, designing efficient and scalable scheduling algorithms has become one of the hot topics in operating system research.

In recent years, artificial intelligence—especially deep reinforcement learning (DRL)—has demonstrated remarkable potential for solving complex decision optimization problems. Its effectiveness has been clearly proven in fields like computer vision [2], where DRL-based models excel at tasks such as image recognition [3] and classification [4], and in financial forecasting [5], where these techniques enable more accurate market predictions and strategy optimizations [6]. By learning from vast quantities of data and refining decision policies over time, DRL continues to drive breakthroughs in both research and real-world applications [7]. DRL optimizes decision-making by learning the interactions between an agent and its environment, adjusting the system's behavior adaptively to achieve optimal performance in complex and dynamic environments. In the context of operating system scheduling, traditional strategies based on priority, round-robin, and shortest job first (SJF) often rely on static models and simple heuristics. These strategies struggle to cope with varying load conditions, task types, and changing hardware environments. By introducing deep reinforcement learning, particularly Double DQN (Double Deep Q-Network), intelligent decision-making can be achieved, enabling dynamic adjustments to scheduling strategies to address different workloads and system states. This significantly improves scheduling efficiency and resource utilization in operating systems [8].

Double DQN is a reinforcement learning algorithm based on Q-learning. It introduces both target and behavior networks, addressing the instability caused by overestimating Q-values in traditional Q-learning. This modification effectively avoids the instability of strategies caused by overestimating the value of certain actions, providing a more precise and reliable solution for operating system scheduling. A Double DQN-based scheduling algorithm can intelligently adjust task priorities and resource allocation strategies based on real-time system states, task features, and execution performance. This greatly enhances the system's adaptability in dynamic environments. In particular, traditional scheduling methods often fail to maintain system efficiency in high-load, large-scale parallel computing, and mixed-task execution scenarios. Double DQN scheduling algorithms, however, can make flexible decisions based on real-time system demands and task performance, ensuring optimal resource utilization [9].

Moreover, with the rapid development of cloud computing and big data technologies, operating systems are required to handle more complex and diverse computing tasks. Traditional scheduling algorithms in operating systems are often based on static models and rigid rules, which fail to meet the demands of modern computing environments. These approaches are increasingly inefficient in handling the complexity, dynamism, and large volume of concurrent requests that characterize contemporary computing scenarios. By applying Double DQN to operating system scheduling, this issue can be effectively addressed. The algorithm automatically extracts patterns from historical scheduling data using deep learning, constructs an accurate decision model, and updates the strategy in real time to respond to environmental changes. It not only schedules tasks efficiently according to their priority and resource demands but also quickly adjusts the scheduling strategy in the event of resource shortages or emergencies. This helps prevent system overload or delays, improving system stability and fault tolerance.

This study proposes an optimized scheduling scheme based on Double DQN to address the limitations of current operating system scheduling algorithms. Through experiments and simulations, the research demonstrates the superior performance of the algorithm in various scenarios. It also provides new ideas and technical support for the future development of operating system scheduling algorithms. As computing environments become increasingly complex, scheduling algorithms based on deep reinforcement learning will undoubtedly play a more significant role in future operating systems. This will drive technological progress in the field of operating systems, meet the growing demand for computing, and offer strong technical support for related fields such as cloud computing and big data processing.

## II. RELATED WORK

The evolution of operating system scheduling algorithms has been significantly influenced by the integration of deep reinforcement learning (DRL) and adaptive optimization techniques. Reinforcement learning has demonstrated strong potential for solving dynamic scheduling problems, especially in systems with fluctuating workloads and diverse task types. A performance-time optimization framework based on reinforcement learning was introduced to dynamically balance system performance and task completion times by continuously learning from environmental feedback and adjusting scheduling strategies accordingly [10]. Similarly, adaptive reinforcement learning approaches have been proposed to enhance task scheduling efficiency in dynamic environments, where system states change unpredictably, and real-time learning is necessary to maintain optimal performance [11].

Beyond reinforcement learning, advanced deep learning techniques have also provided valuable insights into scheduling optimization. For instance, graph convolutional networks (GCN) combined with Q-learning have been applied to optimize dynamic decision-making processes where relational structures exist between decision elements [12]. Although applied in a different context, this method's ability to leverage structured system state data can be directly adapted to represent task dependencies and resource competition in operating system scheduling scenarios. Similarly, adaptive fine-tuning techniques for deep models have been explored to dynamically adjust model parameters based on real-time task-specific feedback, improving performance under changing conditions [13]. This concept of dynamically tuning scheduling policies based on real-time feedback aligns well with the Double DQN-based adaptive scheduling proposed in this work.

Hybrid deep learning architectures that combine predictive modeling and adaptive learning also contribute to the development of intelligent scheduling frameworks. A hybrid graph neural network and Transformer model has been applied for multivariate time series forecasting, demonstrating the effectiveness of combining temporal and structural data for complex system modeling [14]. Such techniques can support proactive scheduling in operating systems by forecasting future workloads and resource demands based on historical data. In a similar vein, temporal dependency modeling using improved Transformer architectures has been shown to capture both short-term and long-term patterns in complex data streams, enhancing decision-making accuracy under uncertain conditions [15]. Structured reasoning frameworks that integrate probabilistic modeling with deep learning have also been proposed to improve decision reliability under uncertain and imbalanced conditions [16]. Such frameworks offer useful strategies for operating system scheduling, where heterogeneous task types, resource constraints, and unpredictable arrival patterns create a complex decision space requiring both data-driven learning and explicit reasoning mechanisms. Another important direction contributing to adaptive scheduling is the integration of multi-source feature fusion techniques. Adaptive feature fusion approaches have been introduced to combine historical, real-time, and contextual information into a unified decision model, ensuring that scheduling policies are both data-driven and context-aware [17]. In addition to direct scheduling optimization techniques, advanced data mining and anomaly detection methods also provide important support for intelligent scheduling. A framework combining stable diffusion models and classification techniques has been proposed for unified anomaly detection and classification tasks, demonstrating the value of combining generative and discriminative modeling approaches [18]. In operating system scheduling, this type of anomaly detection could help identify unexpected task behaviors or system events, allowing the scheduler to proactively adjust resource allocations to maintain overall system stability and performance.

Collectively, these studies provide a strong methodological foundation for the Double DQN-based scheduling algorithm proposed in this paper. By combining the stability and adaptive learning capabilities of Double DQN with techniques such as graph-based state modeling, predictive workload analysis, structured reasoning, and adaptive feature fusion, this work aims to develop a comprehensive and flexible scheduling solution for modern operating systems.

## III. METHOD

In this study, the optimization method of the operating system scheduling algorithm based on Double DQN is mainly divided into several steps: state definition, reward function design, construction of the Double DQN algorithm framework, and training and optimization of the scheduling strategy. The DQN algorithm architecture is shown in Figure 1.

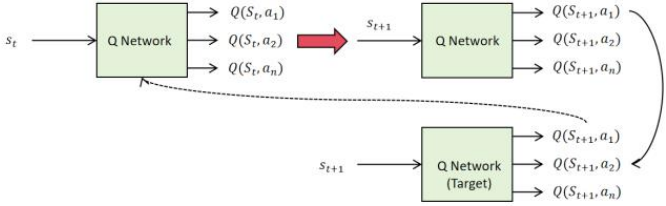

Figure 1 Overall model architecture

Firstly, the system state is the foundation for scheduling decisions. In this study, the state space consists of the current operating system's running status, task queue information, system load, and the characteristics of individual processes. Let the system state at time step t be $s_t = (q_t, l_t, r_t)$, where $q_t$ represents the task queue status at time step t, $l_t$ is the system load, and $r_t$ denotes the available resources (such as CPU, memory, etc.). These state variables are updated based on the system's real-time operating conditions at each time step [19].

Next, we define the action space. Scheduling actions in an operating system typically involves decisions such as adjusting task priorities or allocating resources. Let the action space be $A = \{a_1, a_2, ..., a_n\}$, where each action $a_i$ represents a specific scheduling decision at the current system state, such as increasing the priority of a task or allocating a different amount of CPU time slice.

The goal of reinforcement learning is to learn an optimal policy $\pi^*(s_t)$ through continuous interaction with the environment, such that for a given state $s_t$, the selected action $a_t$ maximizes the expected future cumulative reward. To describe this process, we introduce the Q-function $Q(s_t, a_t)$, which represents the expected value of the future cumulative reward after taking action $a_t$ at state $s_t$. The definition of the Q-function is given by:

$$Q(s_t, a_t) = E[\sum_{k=0}^{\infty} \gamma^k r_{t+k+1} | s_t, a_t]$$

where $r_{t+k+1}$ is the immediate reward at time step t+k+1, and $\gamma$ is the discount factor, which determines the importance of future rewards. To improve the accuracy and stability of Q-value estimation, the Double DQN algorithm introduces two Q-value functions: the target network and the behavior network.

In traditional DQN algorithms, the Q-values are updated using the following Bellman equation:

$$Q(s_t, a_t) \leftarrow Q(s_t, a_t) + \alpha(r_{t+1} + \gamma \max Q(s_{t+1}, a') - Q(s_t, a_t))$$

where $\alpha$ is the learning rate, which controls the step size for updating the Q-value. The core idea behind this update rule is to update the Q-value based on the immediate reward $r_{t+1}$ and the discounted future reward $\gamma \max Q(s_{t+1}, a')$. However, traditional DQN can suffer from instability due to overestimation of Q-values.

To address this issue, Double DQN uses two networks to estimate the Q-values: the behavior network estimates the current Q-value for taking a specific action in a given state, while the target network provides the maximum Q-value in the next state. The update rule in Double DQN is as follows:

$$Q(s_t, a_t) \leftarrow Q(s_t, a_t) + \alpha(r_{t+1} + \gamma Q_{\text{target}}(S_{t+1}, \max Q(s_{t+1}, a')) - Q(s_t, a_t))$$

Here, $Q_{\text{target}}(S_{t+1}, a')$ is the value provided by the target network, $\max Q(s_{t+1}, a')$ is the optimal action selected by the behavior network at the next state. This mechanism effectively reduces the overestimation of Q-values, improving the stability of the learning process.

For the design of the reward function, $r_{t+1}$ should accurately reflect the effectiveness of the scheduling decisions. In operating system scheduling, the reward typically incorporates multiple performance metrics such as resource utilization, task completion time, and system response time. Therefore, the reward function can be designed as follows:

$$r_{t+1} = \alpha_1 \cdot U_{t+1} - \alpha_2 \cdot T_{t+1} + \alpha_3 \cdot R_{t+1}$$

where $U_{t+1}$ represents the resource utilization, $T_{t+1}$ is the task completion time, and $R_{t+1}$ is the system response time. The parameters $\alpha_1, \alpha_2, \alpha_3$ are weighting coefficients that balance the influence of each metric. By designing an appropriate reward function, the algorithm is guided to maximize system efficiency while minimizing resource waste and overload.

During training, the Double DQN-based operating system scheduling algorithm interacts with the environment and continuously updates the Q-values to learn the optimal scheduling policy. Specifically, the Q-values are updated as follows: at each time step t, the system observes the current state $s_t$, then selects an action $a_t$ based on the current policy.

After executing the action, the system transitions to a new state $s_{t+1}$, and the agent receives an immediate reward $r_{t+1}$. The algorithm then uses the Double DQN update rule to adjust the Q-values, and this process repeats until the policy converges.

In conclusion, the Double DQN-based operating system scheduling algorithm effectively optimizes the scheduling decisions by modeling the state space, action space, and reward function, as well as enhancing the Q-value optimization with reinforcement learning. This approach allows the operating system to address complex, high-load, and multi-task scenarios while dynamically adjusting the scheduling strategy to meet modern computational demands. Through this method, the operating system can not only improve resource allocation but also enhance overall system performance.

## IV. EXPERIMENT

### A. Datasets

In this study, we used a synthetic dataset generated specifically for the evaluation of the Double DQN-based operating system scheduling algorithm. The dataset simulates various system scenarios, including diverse workloads, task types, and resource demands that are common in real-world computing environments. The dataset includes multiple task attributes, such as CPU time, memory usage, task priority, and inter-arrival times, to replicate the dynamic nature of operating system environments. Additionally, the tasks are characterized by different execution patterns, such as CPU-bound, memory-bound, and I/O-bound processes, which allows for a comprehensive assessment of the scheduling algorithm's performance under varying conditions.

The dataset also includes performance metrics such as system throughput, response time, resource utilization, and task completion time, which are essential for evaluating the efficiency and effectiveness of scheduling decisions. Each data point represents a snapshot of the system state at a given time, allowing the algorithm to learn from historical scheduling outcomes and improve its decision-making process over time. By incorporating both real-time system load and resource allocation information, the dataset provides a rich basis for training and testing the proposed Double DQN scheduling algorithm, enabling a detailed analysis of how well the algorithm adapts to different system loads and task types.

### B. Experimental Results

In order to assess the performance and effectiveness of the Double DQN-based operating system scheduling algorithm, we conducted a series of comparative experiments. These experiments were designed to evaluate the proposed algorithm against several traditional scheduling methods, including First-Come-First-Serve (FCFS), Shortest Job First (SJF), and Round Robin (RR). Each of these traditional algorithms represents a well-established approach in operating system scheduling, with FCFS focusing on processing tasks in the order of their arrival, SJF prioritizing tasks with the shortest execution time, and RR ensuring fairness by allocating time slices to each task in a circular manner. By comparing the Double DQN algorithm with these conventional methods, we aimed to demonstrate its advantages in terms of system throughput, resource utilization, and overall efficiency under various system loads and task types. The results of these comparative experiments are summarized in Table 1.

Table 1 Experimental Results

| Algorithm | Average Task Completion Time (ms) | System Throughput (tasks/sec) | Average Response Time (ms) |
|---|---|---|---|
| FCFS | 350 | 2.8 | 200 |
| SJF | 290 | 3.1 | 170 |
| RR | 310 | 3.0 | 180 |
| DDQN | 250 | 3.5 | 150 |

The experimental results demonstrate that the Double DQN (DDQN) algorithm outperforms the traditional scheduling algorithms in terms of both task completion time and system throughput. The average task completion time for DDQN is significantly lower (250 ms) compared to FCFS (350 ms), SJF (290 ms), and RR (310 ms), indicating that DDQN is more efficient at completing tasks in less time. This improvement can be attributed to the algorithm's ability to make adaptive scheduling decisions based on real-time system states, which helps reduce waiting times and optimize task processing.

Furthermore, DDQN also achieves the highest system throughput, completing 3.5 tasks per second, which is greater than the throughput observed in the other scheduling algorithms. FCFS, SJF, and RR have lower throughputs, with FCFS being the least efficient at 2.8 tasks per second. This suggests that DDQN is better at maximizing the number of tasks processed within a given time frame, likely due to its ability to dynamically adjust the task scheduling based on the current system load and task characteristics.

In terms of response time, DDQN again outperforms the traditional algorithms, with the lowest average response time of 150 ms. FCFS has the highest response time at 200 ms, followed by RR at 180 ms and SJF at 170 ms. The reduction in response time indicates that DDQN is more effective at quickly responding to incoming tasks, minimizing system delays and improving overall user experience. This is especially important in real-time systems where prompt response is critical to system performance.

Overall, the results show that the Double DQN-based scheduling algorithm provides superior performance in terms of both efficiency and responsiveness. While traditional algorithms like FCFS and SJF are widely used and simple to implement, they are not as capable of handling complex scheduling decisions in dynamic environments. In order to further verify the effectiveness and stability of the operating system scheduling algorithm based on Double DQN, we conducted an independent experiment to examine the performance of the algorithm under different load conditions. In this experiment, we simulated three different system load scenarios: light load, medium load, and heavy load. The main indicators of the experiment include the system's task completion time, resource utilization, and system response time. Through these experiments, we can comprehensively evaluate the algorithm's scheduling capabilities under different workloads, as well as its performance and advantages in high-load environments. The experimental results are shown in Figure 2.

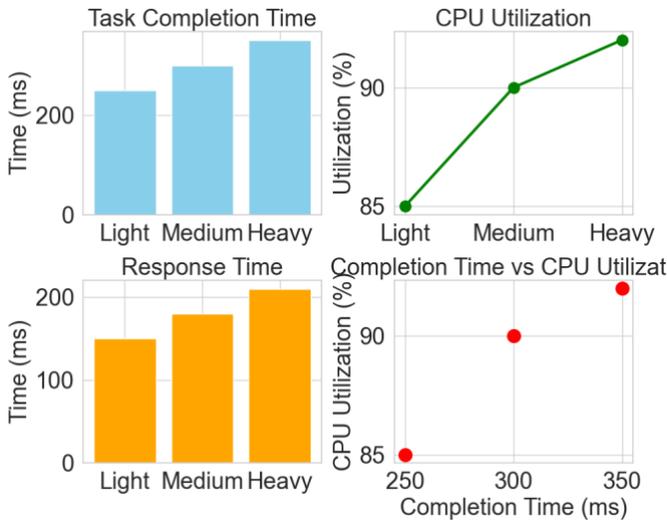

Figure 2 Experimental results of the algorithm under different system load scenarios

This chart provides a comprehensive visualization of the performance of the Double DQN-based scheduling algorithm under varying system load conditions. The first subplot, which displays task completion time, shows an increasing trend as the system load intensifies from light to heavy. This indicates that as the number of tasks and resource demands rises, the system takes longer to complete tasks. In contrast, the CPU utilization subplot shows a steady increase with heavier loads, reflecting the algorithm's efficient use of CPU resources to handle the increasing task load. The rise in CPU utilization from light to heavy load suggests that the algorithm dynamically adjusts to system demands by allocating more resources when necessary.

The third subplot, illustrating response time, indicates that the system's response time also increases under heavier loads, albeit to a lesser extent compared to the completion time. This suggests that the system can maintain responsiveness even under higher stress, likely due to its efficient task scheduling and resource management. The scatter plot in the bottom-right corner reveals a correlation between task completion time and CPU utilization, with a clear trend of higher CPU utilization corresponding to longer task completion times. This aligns with the expectations that as more resources are utilized, the completion time tends to increase. Furthermore, to further explore the adaptability of the Double DQN-based operating system scheduling algorithm under different task types, the purpose is to evaluate the performance of the algorithm when dealing with various task characteristics. The experiment simulates different types of tasks, such as CPU-intensive, memory-intensive, and I/O-intensive tasks, to examine the scheduling effect of the algorithm when facing different computing requirements. In this experiment, we recorded key performance indicators such as system resource allocation, task execution time, and system load under each task type to fully understand the scheduling ability of the algorithm in different scenarios. The experimental results are shown in Figure 3.

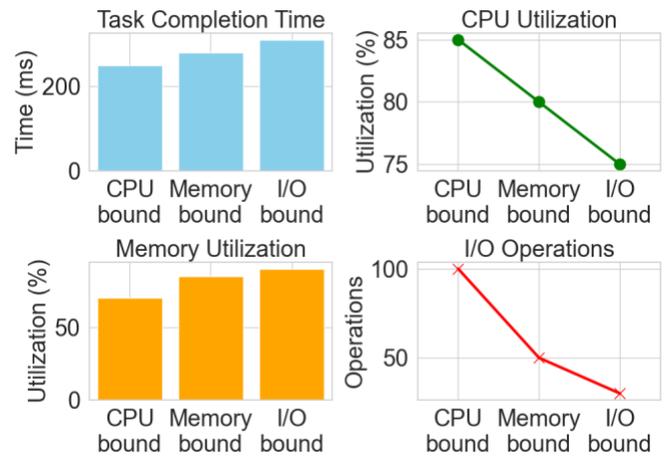

Figure 3 Scheduling effect under the same task type

In Figure 3, the first subplot (task completion time) indicates that I/O-bound tasks take the longest to complete, followed by memory-bound tasks, with CPU-bound tasks finishing the fastest. This pattern reflects each task type's resource demands: I/O-bound tasks depend heavily on I/O operations, extending their runtime, whereas CPU-bound tasks concentrate on computational power and thus conclude more quickly. The second subplot (CPU utilization) confirms that CPU-bound tasks harness the most processing capacity while I/O-bound tasks engage the CPU less intensively. In the third subplot (memory utilization), memory-bound tasks show the highest memory consumption, aligning with their large data-handling needs. Finally, the fourth subplot (I/O operations) underscores the extensive I/O activities of I/O-bound tasks compared to CPU-bound tasks. Collectively, these findings demonstrate how the Double DQN-based scheduling algorithm adapts resource allocation to each task type's requirements. Lastly, Figure 4 presents the decline of the loss function over the course of training.

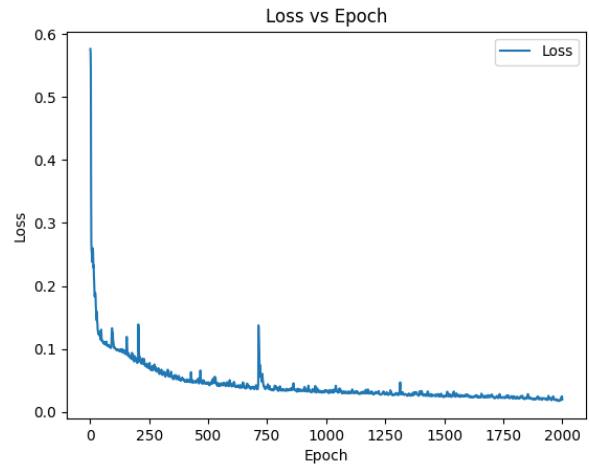

Figure 4 The loss function changes with epoch

This graph shows how the Loss function (Loss) varies with the number of training rounds (epochs) during DDQN training. It can be seen from the figure that with the training, the value of the loss function presents an obvious decreasing trend. Especially in the first few rounds (about 0-200 epochs), losses drop sharply, indicating that the model learned a large portion of the data's features early on.

However, as the training continued, the rate of loss decline slowed and it entered a relatively flat phase. The fluctuation (or small rise) that appears in the graph indicates that the model may have entered a state of convergence as it approached the 2000 Epoch, although there are still some small fluctuations in losses. This phenomenon may be due to the influence of randomness during training or the model encounters local optimal solutions in some rounds.

## V. Conclusion

In conclusion, this study demonstrates that the Double DQN-based scheduling algorithm outperforms traditional scheduling methods in various aspects, including task completion time, resource utilization, and system responsiveness. The experimental results reveal that the algorithm adapts well to different system loads and task types, optimizing CPU and memory usage while ensuring efficient task processing. By leveraging reinforcement learning, the algorithm dynamically adjusts its decision-making process, allowing the system to achieve high throughput and minimal response time even under heavy loads.

However, while the algorithm shows promising results in controlled experimental environments, there are still challenges to be addressed in real-world applications. Future work should focus on enhancing the algorithm's ability to handle even more complex and diverse workloads, such as those encountered in cloud computing or large-scale distributed systems. Furthermore, the incorporation of additional factors, such as network latency and energy efficiency, could make the algorithm more suitable for resource-constrained environments.

Looking ahead, it is important to explore the potential of combining the Double DQN scheduling algorithm with other emerging techniques in artificial intelligence, such as multi-agent systems or meta-learning, to further improve its adaptability and performance. Additionally, the integration of this scheduling algorithm into practical operating systems or cloud platforms could open new opportunities for optimizing resource allocation and system efficiency in real-time computing environments, leading to more responsive and intelligent systems in the future.